\documentclass[10pt,twocolumn,letterpaper]{article}

\usepackage[pagenumbers]{cvpr} 

\usepackage{multirow}
\usepackage{makecell}
\usepackage{bbding}
\usepackage{bbm}
\usepackage{amssymb}
\usepackage{colortbl}

\usepackage{xcolor}
\definecolor{SpringGreen4}{RGB}{46,139,87}
\definecolor{Honeydew4}{RGB}{131, 139, 131} 

\definecolor{cvprblue}{rgb}{0.21,0.49,0.74}

\usepackage[colorlinks, linkcolor=red,  anchorcolor=blue, citecolor=cvprblue]{hyperref}

\def\paperID{*****} 
\def\confName{ICCV}
\def\confYear{2025}

\title{ RGB-Event based Pedestrian Attribute Recognition: A Benchmark Dataset and An Asymmetric RWKV Fusion Framework }

\author{
Xiao Wang$^{1}$, Haiyang Wang$^{1}$, Shiao Wang$^{1}$, Qiang Chen$^{1}$, Jiandong Jin$^{2}$, \\ 
Haoyu Song$^{1}$, Bo Jiang$^{1}$\thanks{Corresponding Author: Bo Jiang}, Chenglong Li$^{2}$ \\ 
${^1}$School of Computer Science and Technology, Anhui University, Hefei, China \\
${^2}$School of Artificial Intelligence, Anhui University, Hefei, China \\ 
\textit{why2434961256@163.com}, \textit{\{xiaowang, jiangbo\}@ahu.edu.cn}, \textit{wsa1943230570@126.com}, \\ 
\textit{\{e23301220, e22214005\}@stu.ahu.edu.cn}, \textit{\{jdjinahu, lcl1314\}@foxmail.com}
}

\begin{document}
\maketitle
\begin{abstract}
Existing pedestrian attribute recognition methods are generally developed based on RGB frame cameras. However, these approaches are constrained by the limitations of RGB cameras, such as sensitivity to lighting conditions and motion blur, which hinder their performance. Furthermore, current attribute recognition primarily focuses on analyzing pedestrians' external appearance and clothing, lacking an exploration of emotional dimensions. In this paper, we revisit these issues and propose a novel multi-modal RGB-Event attribute recognition task by drawing inspiration from the advantages of event cameras in low-light, high-speed, and low-power consumption. Specifically, we introduce the first large-scale multi-modal pedestrian attribute recognition dataset, termed EventPAR, comprising 100K paired RGB-Event samples that cover 50 attributes related to both appearance and six human emotions, diverse scenes, and various seasons. By retraining and evaluating mainstream PAR models on this dataset, we establish a comprehensive benchmark and provide a solid foundation for future research in terms of data and algorithmic baselines. In addition, we propose a novel RWKV-based multi-modal pedestrian attribute recognition framework, featuring an RWKV visual encoder and an asymmetric RWKV fusion module. Extensive experiments are conducted on our proposed dataset as well as two simulated datasets (MARS-Attribute and DukeMTMC-VID-Attribute), achieving state-of-the-art results. The source code and dataset will be released on \textcolor{red}{\url{https://github.com/Event-AHU/OpenPAR}} 
\end{abstract}

\section{Introduction} \label{Introduction} 

Existing Pedestrian Attribute Recognition (PAR)~\cite{wang2022pedestrian} systems are typically developed using RGB frame cameras, with the goal of enabling a model to accurately select matching attributes from a predefined list based on a given pedestrian image or video. Common attributes include \textit{appearance information} such as hairstyle, gender, upper and lower body clothing, and carried objects. With the advancement of artificial intelligence technology, pedestrian attribute recognition has garnered increasing attention and is now widely applied in various human-centric visual tasks, including pedestrian detection and tracking~\cite{wang2022probabilistic, wang2023uni3detr, li2022unsupervised, liu2023uncertainty}, person re-identification~\cite{wang2022uncertainty, hong2021fine, yu2020cocas}, and text-image retrieval~\cite{galiyawala2021person, yang2023towards, zhu2021dssl} for pedestrians. Despite significant progress, the overall performance of these systems remains constrained by the inherent limitations of RGB cameras, such as sensitivity to lighting conditions, clutter background, and susceptibility to motion blur. Interestingly, while multi-modal fusion has been extensively adopted in other computer vision tasks~\cite{xue2024bi, xu2024dmr, shan2024contrastive}, it is surprising that pedestrian attribute recognition still predominantly relies on single-modality RGB cameras. This limitation highlights an opportunity for innovation through the integration of additional modalities to address the existing performance bottlenecks. 
\begin{figure}
    \centering
    \includegraphics[width=1\linewidth]{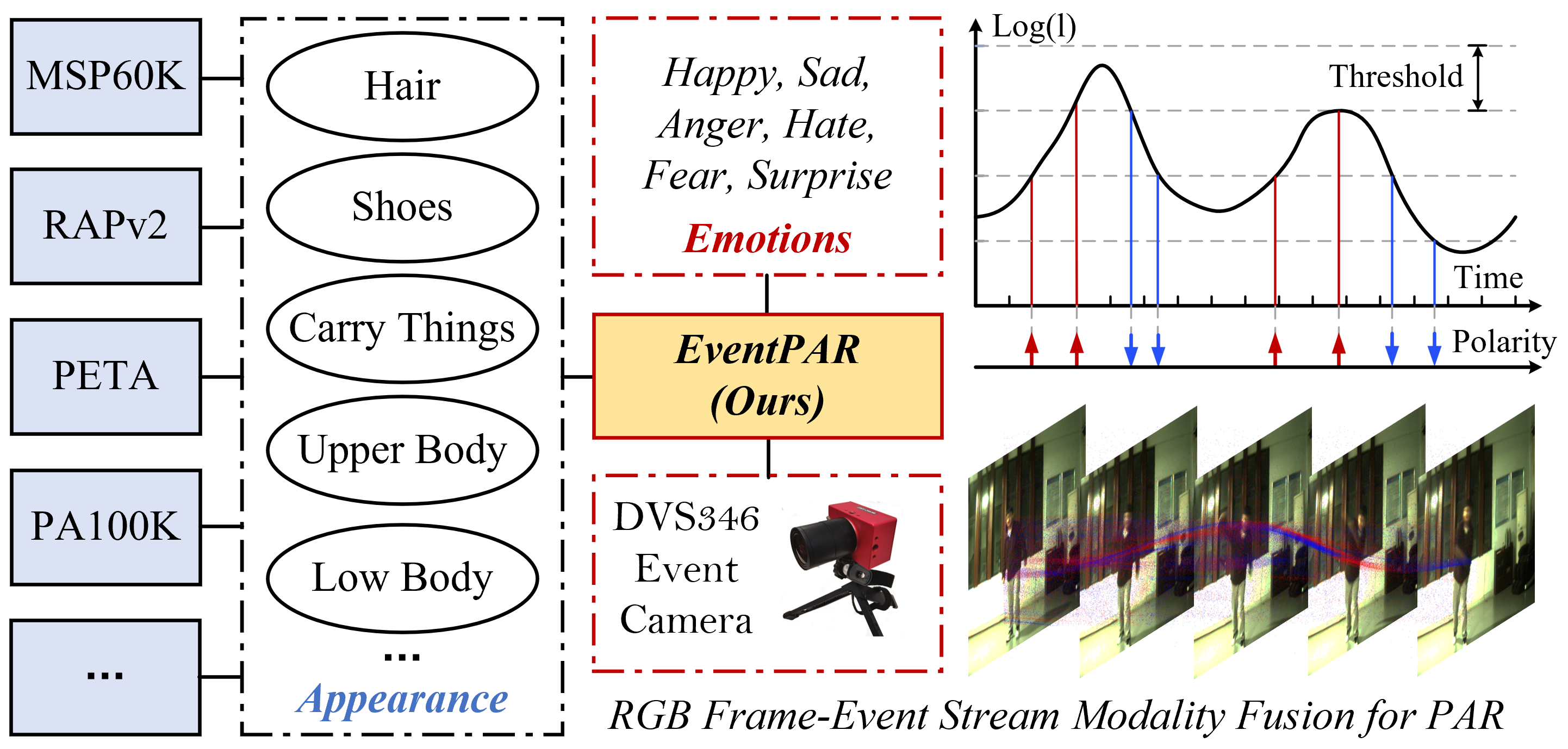}
    \caption{(left) Comparison between existing PAR datasets and our newly proposed EventPAR; (right) Imaging principle and results of Frame and Event cameras.} 
    \label{fig:Eventcameras}
\end{figure}

On the other hand, the pedestrian attributes defined in existing PAR benchmark datasets focus on describing the visible appearance cues, but none of them consider the invisible \textit{emotion information}, which is also very important for human-centric perception. Recognizing emotions such as anxiety, anger, or joy can help the system better understand pedestrian behavioral intent and enhance the naturalness of human-computer interaction, improve safety and risk assessment, and support personalized services. Incorporating emotion analysis into pedestrian attribute recognition tasks can make the system smarter and more human-centric, while also providing support for a broader range of practical application scenarios. 

In this work, we first propose to recognize human attributes using a DVS346 event camera that outputs aligned frame-event streams. This is because event cameras have inherent advantages in high dynamic range, high temporal resolution, and low power consumption. However, event streams contain some noise and are less sensitive to static objects. Therefore, combining them with RGB frame cameras is necessary to achieve more robust and accurate results. Due to the lack of datasets in this area, we first propose a large-scale multi-modal RGB-Event pedestrian attribute recognition benchmark dataset, termed \textbf{EventPAR}. This dataset covers common human appearance attributes as well as six human emotional attributes, totaling 50 attributes and 100K paired RGB-Event samples, as shown in Fig.~\ref{fig:Eventcameras}. These videos were collected under various challenging scenarios, including different lighting conditions, motion blur, object occlusion, and adverse weather conditions. In addition to natural challenging factors, we also artificially introduced some adversarial attack noises and degradation strategies, such as various types of noise and occlusion. To provide a solid benchmark for future research, we re-train and report 17 mainstream PAR algorithms, as shown in Table~\ref{resultsEventPAR}. 

Based on the proposed EventPAR dataset, we introduce a novel multi-modal RGB-Event pedestrian attribute recognition framework leveraging the RWKV model~\cite{duan2024vision}, which effectively integrates event streams to further enhance pedestrian attribute recognition results from the traditional RGB frame. Specifically, given the input RGB image and event streams, we first employ an RWKV encoder to extract visual features from both modalities. These features are then fed into an asymmetric RWKV fusion module, designed to better fuse spatial features from RGB frames with temporal information from event streams. This fusion module efficiently removes redundant event tokens while achieving interactive fusion with RGB features. Finally, a linear layer maps the fused features to multi-label attribute predictions. We extensively validate the proposed framework on three datasets and achieve state-of-the-art performance. 

To sum up, we draw the main contributions of this paper as the following three aspects: 

1). We propose to recognize the pedestrian attributes using an event camera and construct a large-scale benchmark dataset termed EventPAR. It contains 100K aligned RGB frames and event streams, collected from multiple scenarios, illuminations, etc. 

2). We propose a novel RWKV-based framework for frame-event fusion based attribute recognition. A novel asymmetric RWKV fusion module fuses the RGB frame and event streams effectively and efficiently. 

3). We build a large-scale benchmark for the EventPAR dataset by re-training 17 representative and strong PAR algorithms. This benchmark will be a good platform to boost the development of event-based attribute recognition further. 

In addition, we also conduct extensive experiments on three PAR benchmark datasets, including the newly proposed EventPAR, MARS-Attribute, and DukeMTMC-VID-Attribute datasets. More experiments fully validated the effectiveness of our proposed model for the frame-event pedestrian attribute recognition.

\section{Related Works} \label{RelatedWorks}

\subsection{Pedestrian Attribute Recognition} 
Pedestrian Attribute Recognition aims to identify and classify multiple attributes of pedestrians, such as age, gender, clothing type, and accessories, from images or videos. 
For image-based PAR tasks, CLEAR~\cite{bui2024clear} combines pre-trained language models to generate pseudo-descriptions for attribute queries and extracts robust features using Cross Transformers and adapters. HPNet~\cite{liu2017hydraplus} and DAHAR~\cite{wu2020distraction} focus on attribute-related regions via attention mechanisms. PARformer~\cite{fan2023parformer} extracts features by combining global and local perspectives. VTB~\cite{cheng2022simple} introduces a text encoder for pedestrian attribute recognition tasks. 
Visual Prompt Tuning (VPT)~\cite{jia2022visual} and CLIP-Adapter~\cite{gao2024clip} enhance few-shot recognition performance by fine-tuning frozen backbones and combining original CLIP knowledge with limited-shot knowledge, respectively.

Video-based PAR, compared to image-based methods, is a relatively new research topic. Researchers have introduced multi-task models, temporal pooling, and sparse temporal attention modules, among other techniques. Chen et al.~\cite{chen2019temporal} proposed a multi-task model with an attention module to handle each attribute per frame. Specker et al.~\cite{specker2020evaluation} introduced global features to incorporate information from different frames. Li et al.~\cite{lee2021robust} proposed a sparse temporal attention module to select unobstructed frames, enhancing robustness. Thakare et al.~\cite{thakare2024let} calculated the mutual correlations of attribute predictions from pedestrian images across different views, while Liu et al.~\cite{liu2024pedestrian} introduced a spatio-temporal saliency module to capture correlations between attributes in both spatial and temporal domains. 
Different from these works, this paper firstly exploits the multi-modal PAR task using the event camera and proposes a large-scale benchmark dataset and RWKV-based framework for this task.

\subsection{Event-based Vision} 
Event cameras have been widely used in computer vision and multi-modal related tasks. Specifically, in the image restoration and enhancement tasks, EventSR~\cite{wang2020eventsr} proposes a novel end-to-end super-resolution image reconstruction framework that reconstructs low-resolution images from event streams, enhances image quality, and upsamples the enhanced images. 
EG-VSR~\cite{lu2023learning} learns spatial-temporal coordinates and implicit neural representations from RGB frames and event features, leveraging the high temporal resolution of events to achieve video super-resolution at arbitrary resolutions. 
Jiang et al.~\cite{jiang2020learning} and colleagues have introduced a method that effectively restores motion-blurred video sequences from event-based camera data, achieving this by integrating global and local scale visual and temporal knowledge, along with a differentiable directional event filtering module. 
Jiang et al.~\cite{jia2023event} proposed a posterior attention module that adjusts standard attention using the prior knowledge provided by event data, which improves the segmentation performance for moving objects. 
Inspired by these works, in this paper, we attempt to achieve high-performance PAR by fusing the event stream and RGB frames.

\subsection{RWKV Model}
The Receptance Weighted Key Value (RWKV)~\cite{peng2023rwkv} model was first proposed in the field of natural language processing (NLP). Before the emergence of RWKV, Transformers were extensively used in NLP. However, Transformers suffer from quadratic time complexity with respect to sequence length, which significantly limits their efficiency, particularly when processing long sequences. 
Inspired by the linear attention mechanism in the Attention Free Transformer (AFT) \cite{Zhai2021aft}, RWKV introduced the WKV attention mechanism. 
At the same time, the token shift operation was proposed to capture local context. 
This combination allows RWKV to take advantage of both the parallel processing attention mechanism of Transformers and the sequential time-series pattern of Recurrent Neural Networks (RNNs). 
Vision-RWKV \cite{Duan2024vrwkv} then extended the RWKV model to the computer vision field. 
BSBP-RWKV~\cite{zhou2024bsbprwkv} proposes an efficient medical image segmentation framework that integrates Perona-Malik Diffusion for noise suppression with boundary preservation and an RWKV-based architecture to reduce computational complexity. 
RWKV-SAM~\cite{yuan2024rwkvsam} proposes an efficient segment-anything framework with a mixed convolution-RWKV backbone and multiscale token decoder, using a joint training benchmark to achieve twice the inference speed and superior segmentation quality compared to Transformer and vision Mamba baselines. 
Different from these works, this paper first adopts the RWKV model for the pedestrian attribute recognition task.

\begin{figure*}
    \centering
    \includegraphics[width=1\linewidth]{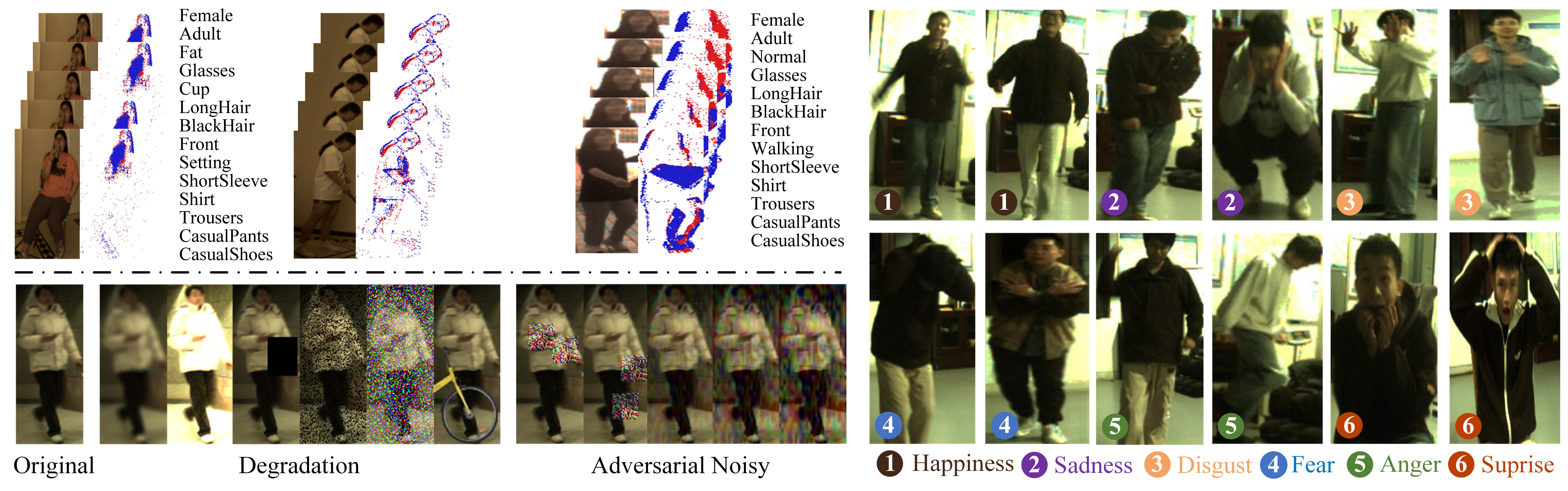}
    \caption{An illustration of representative samples in our newly proposed EventPAR dataset. 
    The left side displays RGB and Event bimodal data collected during summer and winter seasons, as well as the synthetic challenges incorporated into the dataset. The right side presents examples of the six emotion-related attributes.} 
    \label{fig:EventPARsamples}
\end{figure*}

\section{EventPAR Benchmark Dataset} \label{eventPARbenchmark}

\subsection{Protocols} 

In this work, we follow the guidelines below when constructing the EventPAR benchmark dataset: 
\textbf{\textit{1). Multi-Modal:}} 
As shown in Table~\ref{Publicdatasets}, unlike existing frame-based PAR datasets, we utilize the DVS346 camera for data collection, obtaining spatiotemporal aligned multi-modal data comprising visible light and event streams. It is the first multi-modal PAR database, laying a solid data foundation for subsequent research.
\textbf{\textit{2). Emotion Attributes:}} 
Unlike conventional PAR datasets that primarily focus on pedestrians' appearance attributes, our new dataset incorporates six fundamental human emotional attributes, i.e., \textit{Happiness, Sadness, Anger, Surprise, Fear,} and \textit{Disgust}. This addition effectively fills the gap in the attribute recognition domain by addressing the lack of discussion on the emotional analysis dimension. The static and dynamic presentations are shown on the right side of Fig.~\ref{fig:EventPARsamples}. 
\textbf{\textit{3). Diverse Scenes and Cross-Seasonal:}} 
Different from existing PAR datasets that involve short-term data collection, our data acquisition spanned several months, encompassing data from different seasons (summer and winter). Additionally, it covers a variety of scenes and weather conditions, such as daytime and nighttime scenes, as well as weather conditions like sunny and rainy days, as shown in Fig.~\ref{fig:EventPARsamples} (a, b). These elements significantly enhance the diversity and challenge of the dataset, providing better validation of a model's fitting ability and generalization capability. 
\textbf{\textit{4). Large Scale:}} 
It contains 100K spatio-temporally aligned pedestrian samples of RGB frames and event streams, 12 attribute groups, and 50 fine-grained pedestrian attributes. It is one of the largest PAR datasets, comparable to PA100K in scale, as shown in Table~\ref{Attributes}.
\textbf{\textit{5). Simulated Complex Real-world Environments:}} 
Our dataset is carefully designed to incorporate a wide range of challenges, such as variations in illumination, motion blur, object occlusion, and adverse weather conditions, all of which simulate the complex real-world difficulties encountered in pedestrian attribute recognition. To further increase the complexity and practical relevance of the dataset, we employ \textit{adversarial attack} techniques~\cite{carlini2017towards} during data processing, adding an extra layer of difficulty, as shown in Fig.~\ref{fig:EventPARsamples} (c).

\begin{figure}
\centering
\includegraphics[width=0.48\textwidth]{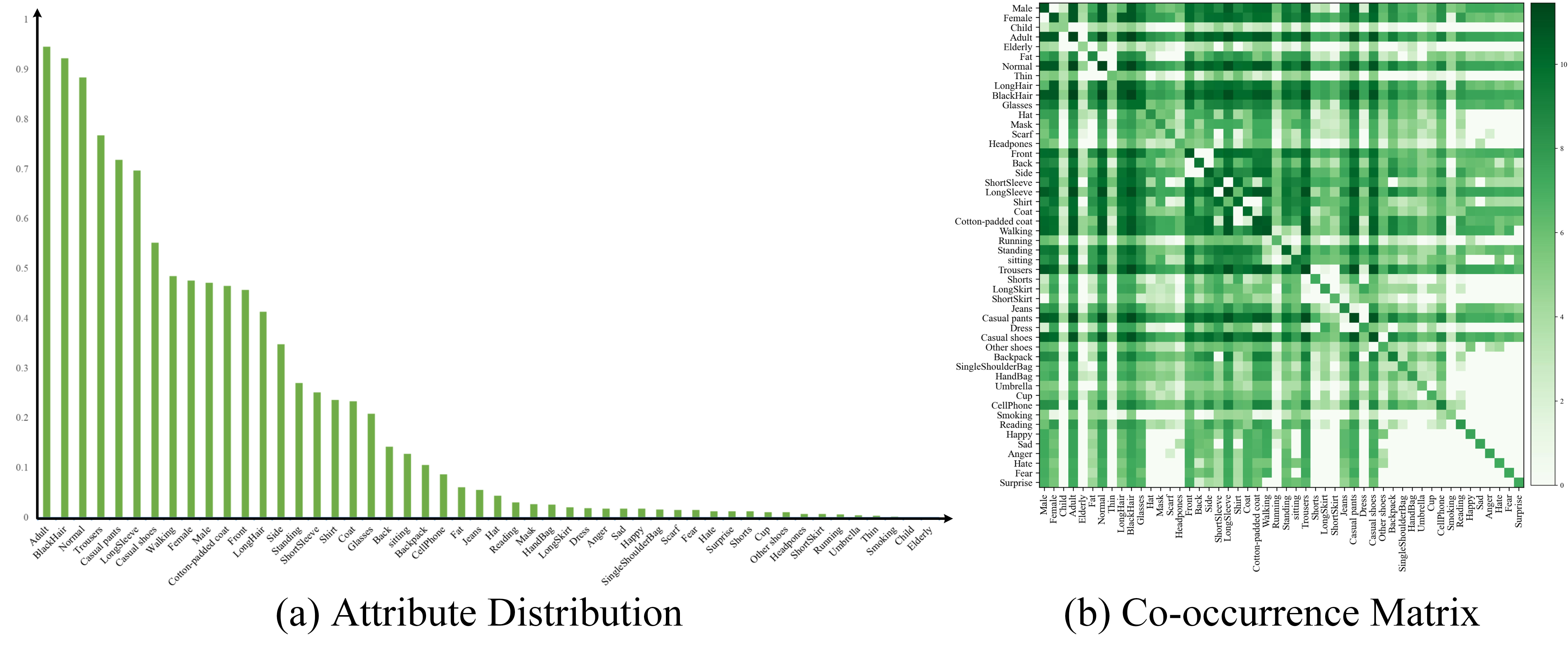}
\caption{(a) Attributes Distribution: Bar graph showing the prevalence of individual attributes across the dataset; (b) Co-occurrence Matrix of Attributes: Logarithmic heatmap showing the co-occurrence frequency of attribute pairs.} 
\label{fig:attrDistribution}
\end{figure}

Moreover, EventPAR exhibits a real-world long-tail distribution, as shown in Fig.~\ref{fig:attrDistribution} (a), where frequent attributes include \textit{Adult, Back Hair,} and \textit{Normal Size} while rare ones such as \textit{Smoking, Child,} and \textit{Elderly} appear less often. The attribute co-occurrence matrix is shown in Fig.~\ref{fig:attrDistribution} (b). 

\begin{table}[t]
    \centering
    \begin{minipage}{\linewidth}
        \centering
        \small 
        \caption{Comparison with existing PAR benchmark datasets. $\#$\textbf{Att.} is short for the number of Attributes. Deg and Adv is short for Degradation and Adversary, respectively.}   
        \label{Publicdatasets} 
        \resizebox{0.99\textwidth}{!}{ 
        \begin{tabular}{l|l|l|l|cccccc}
            \hline \toprule [0.5 pt]
            \textbf{Dataset}  & \textbf{Year} &$\#$\textbf{Att.}  &$\#$\textbf{Samples}  & \textbf{Video}  & \textbf{Frame}  & \textbf{Event}  &\textbf{Emotion}&\textbf{Deg} &\textbf{Adv}  \\ 
            \hline 
            PETA~\cite{deng2014pedestrian} & 2014 & 61 & 19,000  & \XSolidBrush &\Checkmark & \XSolidBrush & \XSolidBrush& 
           \XSolidBrush& 
          \XSolidBrush\\
            WIDER~\cite{li2016human} & 2016 & 14 & 57,524  & \XSolidBrush&\Checkmark  & \XSolidBrush & \XSolidBrush& 
            \XSolidBrush& 
            \XSolidBrush\\
            RAPv1~\cite{li2016richly} & 2016 & 69 & 41,585    & \XSolidBrush &\Checkmark  & \XSolidBrush & \XSolidBrush& 
           \XSolidBrush& 
           \XSolidBrush\\
            PA100K~\cite{liu2017hydraplus} & 2017 & 26 & 100,000 &\XSolidBrush  &\Checkmark  & \XSolidBrush & \XSolidBrush& 
            \XSolidBrush& 
            \XSolidBrush\\
            RAPv2~\cite{li2018richly} & 2019 & 76 & 84,928    & \XSolidBrush  &\Checkmark  & \XSolidBrush & \XSolidBrush& 
            \XSolidBrush& 
            \XSolidBrush\\
            MSP60K~\cite{jin2024pedestrian} & 2024 & 57 & 60,015 & \XSolidBrush  & \Checkmark  & \XSolidBrush & \XSolidBrush& 
            \Checkmark& 
           \XSolidBrush\\
            \hline 
            MARS-Attribute~\cite{zheng2016mars} & 2016 & 43 & 16,360  & \Checkmark  &\XSolidBrush  & \XSolidBrush  & \XSolidBrush& 
            \XSolidBrush& 
           \XSolidBrush\\
            DukeMTMC-VID~\cite{ristani2016performance} & 2017 & 36 & 4,832 & \Checkmark&\XSolidBrush & \XSolidBrush & \XSolidBrush & 
            \XSolidBrush& 
            \XSolidBrush\\
            \hline 
            EventPAR (Ours) & 2025 & 50 & 100,000  & \Checkmark & \Checkmark  &\Checkmark & 
           \Checkmark& \Checkmark&  \Checkmark\\
            \hline \toprule [0.5 pt] 
        \end{tabular} } 
    \end{minipage}

    \vspace{0.5cm} 

    \begin{minipage}{\linewidth}
        \centering
        \caption{Attribute groups and details defined in our proposed EventPAR dataset.} 
        \label{Attributes} 
        \resizebox{0.99\textwidth}{!}{ 
        \begin{tabular}{l|l}
            \hline \toprule [0.5 pt]
            \multicolumn{1}{c|}{\multirow{1}{*}{\textbf{Attribute Group}}} & \multicolumn{1}{c}{\textbf{Details}} \\
            \hline     
            \#01 Gender  &Male, Female \\ 
            \hline    
            \#02 Age  & Child, Adult, Elderly \\ 
            \hline
            \#03 Body Size  & Fat, Normal, Thin \\ 
            \hline
            \#04 Viewpoint  & Front, Back, Side \\ 
            \hline
            \#05 Head  & \makecell[l]{Long Hair, Black Hair, Hat, Glasses, Mask,  Scarf, Headphones} \\ 
            \hline
            \#06 Upper Body  & \makecell[l]{Short Sleeves, Long Sleeves, Shirt, Coat, Cotton-padded Coat} \\ 
            \hline
            \#07 Lower Body  & \makecell[l]{Trousers, Shorts, Jeans, Long Skirt, Short Skirt, Casual Pants, Dress} \\ 
            \hline
            \#08 Shoes  & \makecell[l]{Casual Shoes, Other Shoes} \\ 
            \hline
            \#09 Accessory & \makecell[l]{Backpack, Shoulder Bag, Hand Bag, Umbrella, Cup, Cellphone} \\ 
            \hline
            \#10 Posture  & \makecell[l]{Walking, Running, Standing, Sitting} \\ 
            \hline
            \#11 Activity  & \makecell[l]{Smoking, Reading} \\ 
            \hline
            \#12 Emotion  & \makecell[l]{Happiness, Sadness, Anger, Surprise, Fear, Disgust} \\ 
            \hline \toprule [0.5 pt] 
        \end{tabular} 
        }
    \end{minipage}
\end{table}

\subsection{Benchmark Baselines}   

Based on our newly proposed dataset, we retrained and tested 17 mainstream PAR models, providing a solid baseline for future work on the EventPAR dataset. Specifically, the following algorithms were included: 
\emph{1). CNN-based:} DeepMAR~\cite{deepmar},Zhou et al.~\cite{zhou2023solution}, RethinkPAR~\cite{2021Rethinking}, SSCNet~\cite{2021ssc}, SSPNet~\cite{SHEN2024110194}.
\emph{2). Transformer-based:} DFDT~\cite{ZHENG2023105708}, PARFormer~\cite{fan2023parformer}.
\emph{3). Mamba-based:} MambaPAR~\cite{wang2024SSMSurvey}, MaHDFT~\cite{wang2024empiricalmamba}.
\emph{4). Human-Centric Pre-Training Models for PAR:} PLIP~\cite{zuo2023plip}.
\emph{5). Visual-Language Models for PAR:} VTB~\cite{cheng2022VTB}, Label2Label~\cite{li2022label2label}, PromptPAR~\cite{wang2023pedestrian}, SequencePAR~\cite{jin2023sequencepar}.

\begin{figure*}[!htp]
\centering
\includegraphics[width=0.95\textwidth]{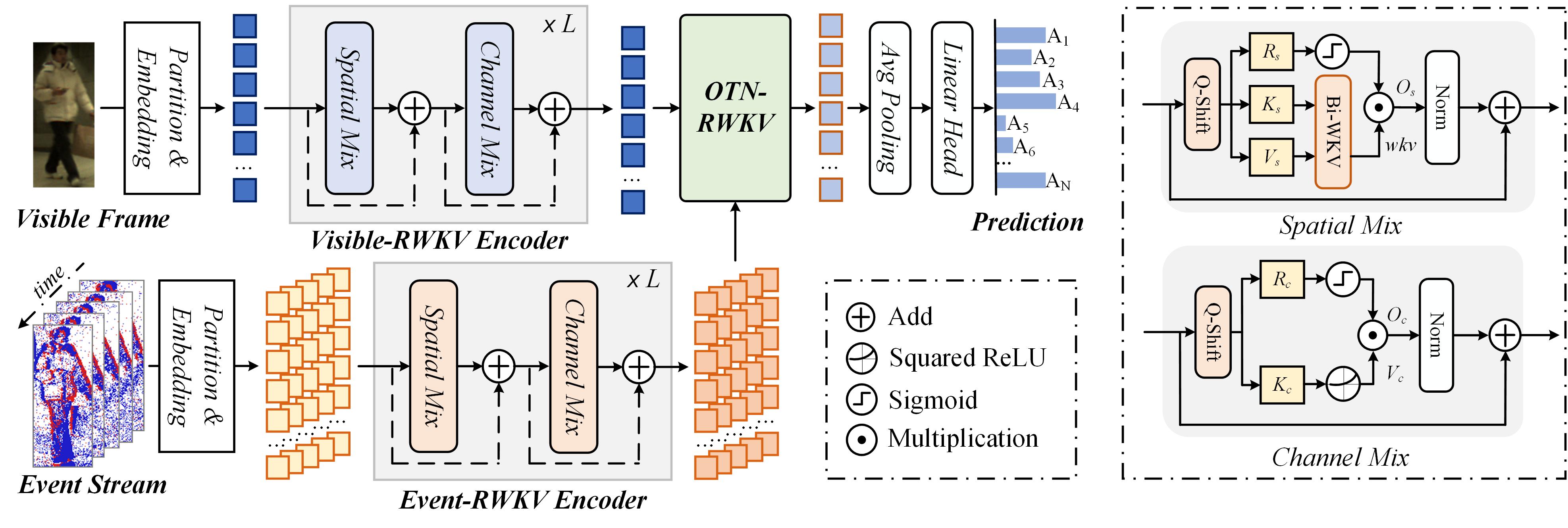}
\caption{An overview of our proposed OTN-RWKV guided RGB-Event fusion for pedestrian attribute recognition. }
\label{fig:framework}
\end{figure*}

\section{Methodology} \label{Methodology} 
\subsection{Overview} 
To improve recognition accuracy and robustness, this paper proposes a video-based multimodal pedestrian attribute recognition framework shown in Fig.~\ref{fig:framework}. Event data is used to supplement RGB information loss, enhancing stability. Features are extracted via Visible-RWKV and Event-RWKV encoders, fused through the OTN-RWKV module, and then classified for attribute prediction. Please refer to the following for more detailed information.

\subsection{Network Architecture} 
\noindent $\bullet$ \textbf{Input Encoding:} 
In this work, event modality is used to assist the RGB video in overcoming information loss in complex environments. 
Given a sequence of an RGB video denote as $X_r\in \mathbb{R}^{T\times H \times W \times C}$, where $T$ is the total number of frames, and $H$, $W$, $C$ correspond to the height, width, and the number of channels of each frame, respectively. We take event streams as $\mathcal{E} = \{e_1, e_2, ..., e_M\}$, each event point $e_j$ can be denoted as $\{x, y, t, p\}$, representing spatial coordinates, timestamps, and polarity, respectively. To align with the RGB sequence, we stack the event streams into event frames that are temporally synchronized with the RGB frames based on exposure times. Therefore, we can obtain the event frame sequence $X_e \in \mathbb{R}^{T\times H \times W \times C}$.

By inputting a sequence of RGB frames $X_r\in \mathbb{R}^{T\times H \times W \times C}$ and a corresponding sequence of event frames $X_e \in \mathbb{R}^{T\times H \times W \times C}$, we first apply a partitioning operation to process data for both modalities, transforming each frame into $ {HW}/p^2 $ patches, where $p$ denotes the patch size. Then, the patches are added with Position Embedding (P.E.), which are used to encodes spatial information to obtain RGB tokens $X_r \in \mathbb{R}^{T\times N \times C}$ and event tokens $X_e\in \mathbb{R}^{T\times N \times C}$ separately, where $N$ is the number of tokens.

\noindent $\bullet$ \textbf{RWKV Encoder:}
In this paper, we employ Vision-RWKV~\cite{duan2024vision} as our visual encoder, which adopts a block-stacked design with $L$ layers to enhance its encoding capability, where each block consists of a Spatial-Mix module and a Channel-Mix module. In each layer, RGB tokens $X_r \in \mathbb{R}^{T\times N \times C}$ and event visual tokens $X_e\in \mathbb{R}^{T\times N \times C}$ are first fed into the Spatial-Mix module, which performs linear complexity global attention computation. Specifically, as shown in the top right of Fig.~\ref{fig:framework}, the input visual tokens first undergo a token Shift operation to capture local context information before being fed into three parallel linear layers to obtain the matrices $R_s$,$K_s$,$V_s\in \mathbb{R}^{T*N \times C}$. The formulas are as follows:
\begin{equation}
\label{spatial}
\begin{aligned}
    R_\text{s} &= {\rm Q\mbox{-}Shift}_R(X_i) W_R,\\ 
    K_\text{s} &= {\rm Q\mbox{-}Shift}_K(X_i) W_K,\\ 
    V_\text{s} &= {\rm Q\mbox{-}Shift}_V(X_i) W_V, \quad i \in (r,e)
\end{aligned}
\end{equation}
where the Q-Shift function is a specialized token shift mechanism specifically designed for vision tasks. After that, $K_s$ and $V_s$ are used to calculate the global attention result $wkv\in \mathbb{R}^{T*N \times C}$
by the Bi-WKV module, which is a linear complexity bidirectional attention mechanism. Subsequently, the $wkv$ is multiplied with $\sigma(R)$, which controls the output $O_i$ probability,
\begin{equation}
\begin{aligned}
O_i &= (\sigma(R_s) \odot wkv) W_O,\\
wkv &= \text{Bi-WKV}(K_s, V_s), \quad i \in (r,e),
\end{aligned}
\end{equation}
where the operator $\sigma$ represents the sigmoid function, while $\odot$ denotes element-wise multiplication. After the output linear projection, the features are subjected to layer normalization to enhance stability and facilitate effective training.

Subsequently, the tokens are fed into the Channel-Mix module to achieve channel-wise fusion, as shown in the bottom right of Fig.~\ref{fig:framework}. The parameters $R_c$ and $K_c$ are computed similarly to the Spatial-Mix before the final output,
\begin{equation}
\begin{aligned}
\label{channel}
R_c &= \text{Q-Shift}_R(O_i) W_R, \\
K_c &= \text{Q-Shift}_K(O_i) W_K, 
\end{aligned}
\end{equation}
the gating mechanism $\sigma(R_c)$ is used to modulate the output $O'_i$. To maintain information integrity and promote gradient flow, residual connections are introduced to fuse the module outputs, thereby effectively alleviating the vanishing gradient problem. This process can be expressed as:
\begin{equation}
\begin{aligned}
O_i' &= (\sigma(R_c) \odot V_c)W_O, \\
V_c &= \text{SquaredReLU}(K_c) W_V,
\end{aligned}
\end{equation}

\begin{figure}
    \centering
    \includegraphics[width=1\linewidth]{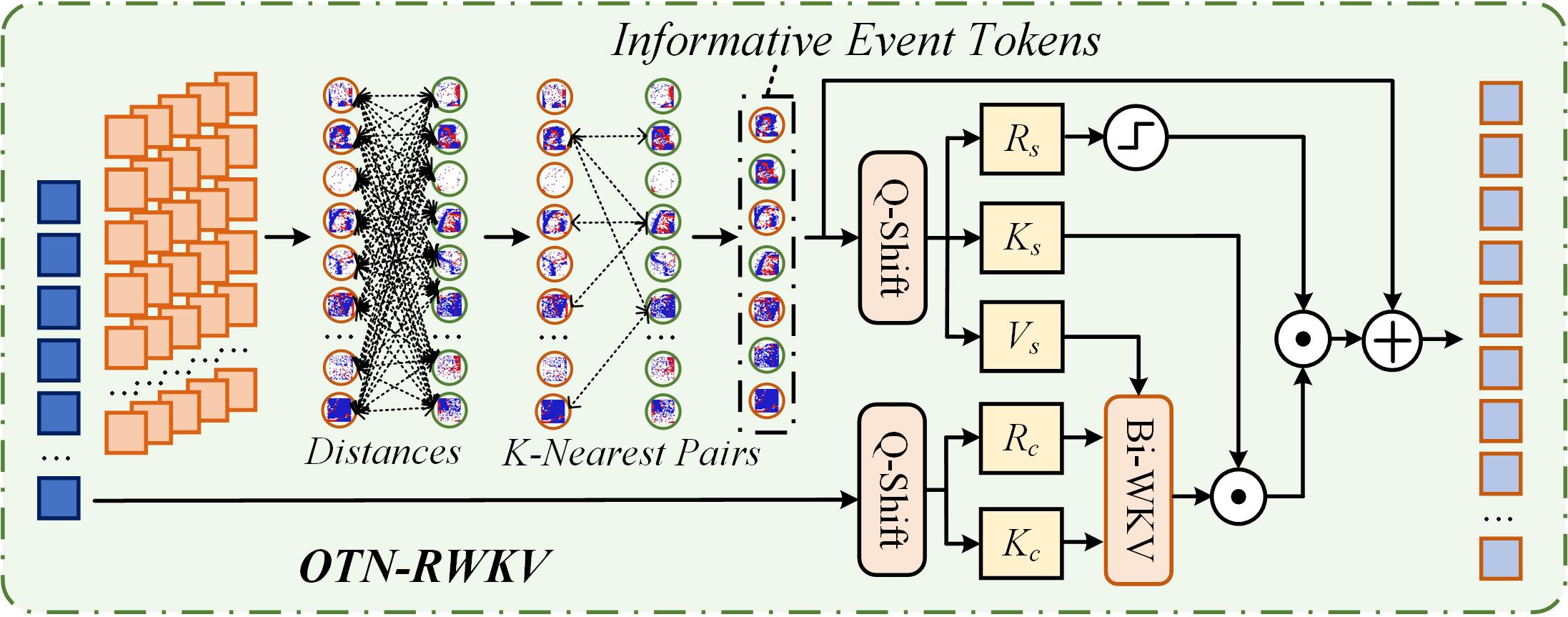}
    \caption{Illustration of our proposed OTN-RWKV for RGB-Event fusion.} 
    \label{fig:OTNRWKV}
\end{figure}

\noindent $\bullet$ \textbf{OTN-RWKV:}
Due to the asynchronous nature of event cameras, event frames contain a large amount of redundant information, increasing computational burden and reducing the effectiveness of auxiliary data. To address this, we propose the OTN-RWKV fusion module, which uses a similarity matrix to identify the top-$K$ most similar token pairs from the event tokens output \( O_e' \), retaining only the most representative tokens. This method improves computational efficiency and enhances the quality of multimodal fusion. The implementation process is as follows:
\begin{equation}
\begin{aligned}
O_{e}^{''} &= \text{KNP}_{filter}(\text{sim}(O_e', O_e')) \odot O_e',  
\end{aligned}
\end{equation}
where $O_e^{''} \in \mathbb{R}^{N' \times C}$ represents the tokens for the informative filtered event. After filtering, the number of tokens in both modalities is aligned to facilitate fusion.

Although traditional fusion methods (e.g., concatenation, addition, and 1×1 convolution) offer strong compatibility, they often cause information loss, limiting the preservation of fine-grained multimodal details. Inspired by the concept of cross attention and LCR\cite{yin2024video}, in this work, we further propose a novel interactive fusion strategy. As shown in Fig.~\ref{fig:OTNRWKV}, we first apply the Q-shift operation separately to the two types of visual tokens, corresponding to Eq.\ref{spatial} and Eq. \ref{channel}, respectively. Subsequently, $V_s$, $R_c$, and $K_c$ are used to compute the cross-attention results through a linear complexity bidirectional attention mechanism,
\begin{equation}
O_{fusion} =  \sigma(R_s) \odot \text{$LN$} (K_s \odot \text{Bi-}WKV) + O_{e}^{'},
\end{equation}
\begin{equation}
\scriptsize
\text{Bi-}WKV(R, K, V)_t = 
\frac{\sum_{i=0, i \neq t}^{M} e^{-(|t-i|-1)/M \cdot v_s + r_i} k_i + e^{u + r_t} k_t}
{\sum_{i=0, i \neq t}^{M} e^{-(|t-i|-1)/M \cdot v_s + r_i} + e^{u + r_t}}.
\end{equation}
where $t$ represents the $t$-th token of the input, $M$ represents the total number of tokens, $u$ is a C-dimensional learnable vector that acts as a bonus indicating the current token, and $LN$ refers to LayerNorm. Unlike LCR\cite{yin2024video}, which uses the learnable vector $w$, we replace $w$  with the output $V_s$ from Eq. \ref{spatial} to better focus on the current sample.

\subsection{Loss Function}  
After the effective fusion of the two modalities using the OTN-RWKV module, the resulting representation is fed into an average pooling layer and then a linear classification head, yielding the final attribute prediction $P_{attr}$. In the training phase, we adopt the weighted cross-entropy loss (WCE Loss)\cite{cheng2022VTB} to alleviate the distribution imbalance between pedestrian attributes, which is widely used by attribute recognition,
\begin{equation}
\small
\mathcal{L}_{wce} = -\frac{1}{N} \sum_{i=1}^{N} \sum_{j=1}^{M} \omega_j \left( y_{ij} \log(p_{ij}) + (1 - y_{ij}) \log(1 - p_{ij}) \right),
\end{equation}
here, $N$ represents the total number of samples, $M$ denotes the number of attributes, and $\omega_j$ is the weight for the $j$-th attribute. $y_{ij}$ and $p_{ij}$ denote the ground truth and predicted result, respectively.

\section{Experiments} \label{Experiments}
\subsection{Datasets \& Evaluation Metric} 
This study benchmarks 17 pedestrian attribute recognition methods and compares our method with state-of-the-art approaches on the benchmark and public datasets MARS-Attribute~\cite{zheng2016mars} and DukeMTMC-VID-Attribute~\cite{ristani2016performance}. 

\noindent $\bullet$ \textbf{MARS-Attribute Dataset} is an annotated dataset for pedestrian attribute recognition, containing multiple multi-label and binary attributes such as pedestrian action, orientation, clothing color, gender, and others. These attributes are decomposed into 43 binary attributes to enhance the performance of pedestrian attribute recognition models for training and testing. The dataset is divided into a training subset and a testing subset, with 8,298 and 8,062 tracklets, respectively. Each tracklet contains approximately 60 frames on average. The dataset includes 625 and 626 distinct pedestrian identities for training and testing.

\noindent $\bullet$ \textbf{DukeMTMC-VID-Attribute Dataset} is an extension of the DukeMTMC-VID dataset, specifically designed for pedestrian attribute recognition. This dataset includes various annotated pedestrian attributes aimed at improving pedestrian re-identification performance. It features 2,032 identities and 16,522 video sequences captured across multiple scenarios, with a focus on real-world and challenging environments. In addition, the dataset provides rich attribute annotations. By splitting the multi-label attributes into binary attributes, the dataset includes a total of 36 binary attributes for training and testing. The training subset contains 702 distinct pedestrian identities and 16,522 images, while the testing subset includes 17,661 images corresponding to 702 pedestrians. The DukeMTMC-VID-Attribute dataset is widely used for evaluating attribute-based pedestrian re-identification models, offering valuable insights into how attributes influence recognition accuracy under varying conditions.

For the evaluation of our and the compared PAR models, we use five widely adopted evaluation metrics to assess performance, including: \textbf{mean Accuracy} (mA), \textbf{Accuracy} (Acc), \textbf{Precision} (Prec), \textbf{Recall}, and \textbf{F1-score} (F1), which can be expressed as:
\begin{equation}
\centering 
\small 
Accuracy=\dfrac{TP+TN}{FP+FN+TP+TN}
\end{equation}
\begin{equation}
\centering 
\small 
Precision=\dfrac{TP}{FP+TP},~~ Recall=\dfrac{TP}{TP+FN}
\end{equation}
\begin{equation}
\centering 
\small 
F1 =\dfrac{2\times Recall\times Precision }{Recall+Precision} 
\end{equation} 
where $TP$ denotes the number of correctly predicted positive samples, $TN$ is the number of correctly predicted negative samples, $FP$ and $FN$ denote the number of false positive and false negative samples, respectively. 

\begin{table*}[!htb]
\center
\setlength{\tabcolsep}{5pt}
\small  
\caption{Comparison with public methods on our datasets.  
Size indicates the size of checkpoint file, and $\#$\textbf{MS} means the GPU memory used for optimization, Test time refers to inference time.}   
\label{resultsEventPAR} 
\resizebox{\textwidth}{!}{ 
\begin{tabular}{l|l|ccccc|l|l|l|l|l}
\hline \toprule [0.5 pt] 
\multirow{1}{*}{\textbf{Methods}} & 
\multirow{1}{*}{\textbf{Publish}} & 
\multicolumn{1}{c}{\textbf{mA}} &
\multicolumn{1}{c}{\textbf{Acc}} & 
\multicolumn{1}{c}{\textbf{Prec}} &
\multicolumn{1}{c}{\textbf{Recall}} &
\multicolumn{1}{c|}{\textbf{F1}} &
\multirow{1}{*}{\textbf{Test Time}} &
\multirow{1}{*}{\textbf{Params} (MB)} & 
\multirow{1}{*}{\textbf{Flops} (GB)} &
\multirow{1}{*}{\textbf{Size} (MB)}  &
\multirow{1}{*}{$\#$\textbf{MS} (MB)}  
\\ \hline 
\#01~~DeepMAR~\cite{deepmar}  & ACPR15  &66.57  &69.53  &74.90  & 88.54   &81.57 &64s&23 &25  & 181&592   \\
\#02~~ALM ~\cite{tang2019improving} & ICCV19   & 57.18 & 64.17 &75.59  & 73.20 &74.38  & 813s&17&8 & 66 &785    \\
\#03~~Strong Baseline~\cite{2021Rethinking} &arXiv20    & 73.75 &61.86  &67.23  &80.78  &75.43  &47s & 24&25 & 91 & 576  \\
\#04~~RethinkingPAR~\cite{2021Rethinking} & arXiv20   & 81.37 & 80.84 & 86.31 & 87.57 & 86.93 &44s &24& 24& 91 & 580 \\
\#05~~SSCNet~\cite{2021ssc} & ICCV21 & 63.10 &66.07  &72.72  & 83.22 & 77.62 &49s &24&9 &90  & 588    \\

\#06~~VTB~\cite{cheng2022VTB}  & TCSVT22    & 88.41 & 83.83 &87.89  &89.31  & 88.53 &243s &93&66 & 333 & 525  \\
\#07~~Label2Label~\cite{li2022label2label} & ECCV22   &72.49  &74.01  &86.60  &79.02  &82.19  &398s & 66&37 & 829 &650    \\
\#08~~DFDT~\cite{ZHENG2023105708} & EAAI22   & 61.71 &63.14  &79.17  &70.63  &74.66  &287s &88& 37&669  &  2584 \\
\#09~~Zhou et al.~\cite{zhou2023solution} & IJCAI23   
& 56.46 & 60.89 &73.37  & 73.62 & 73.50 &42s &234& 24& 536 & 701\\
\#10~~PARFormer~\cite{fan2023parformer} & TCSVT23    &83.12  & 80.48 &85.14  &  88.41&86.53  &438s &195& 205& 756 &2481  \\ 
\#11~~SequencePAR~\cite{jin2023sequencepar} & arXiv23      & 86.27 &84.42  & 88.81 &89.12  &88.83  & 1497s&466& 577& 1776 & 9901 \\
\#12~~VTB-PLIP ~\cite{zuo2023plip} & arXiv23     &67.25  &68.37  & 77.75 &79.72  &78.37  &229s &31&22 &  591&501  \\
\#13~~Rethink-PLIP~\cite{zuo2023plip}  & arXiv23  &   68.75 & 70.03 & 81.82 & 78.04 & 79.89 &37s &21   &23 &144  & 481 \\
\#14~~PromptPAR~\cite{wang2023pedestrian}  &TCSVT24     & 86.51 &82.27  & 86.35 &89.36  & 87.64 &1312s &8&5 & 1200 &2296    \\
\#15~~SSPNet ~\cite{zhou2024pedestrian} & PR24    & 66.92 & 67.49 & 78.73 & 76.90 & 77.80 & 69s &40  &26 & 100  & 1898  \\
\#16~~MambaPAR~\cite{wang2024SSMSurvey} & arXiv24   & 50.01 & 42.32 & 54.81 & 57.31 & 55.63 & 105s &0.64&0.41 & 98 & 802  \\ 
\#17~~MaHDFT~\cite{wang2024empiricalmamba} & arXiv24     & 50.43 & 44.98 & 59.10 & 59.70 & 58.57 & 292s &164&6033 & 499 &1313   \\
\hline

~~~~~~~~~~~~OTN-RWKV (Ours, RGB Only)&-  &79.32&76.00 &82.37  &84.55 &83.22 &68s & 107 & 39& 429&   2200     \\ 

~~~~~~~~~~~~OTN-RWKV (Ours, RGB+Event) &-  &\textbf{87.70}&\textbf{84.94}&\textbf{89.15}&\textbf{89.48}&\textbf{89.18}& 361s  & 201 & 77& 836&   2312    \\ 

\hline \toprule [0.5 pt] 
\end{tabular}} 
\end{table*}
\begin{table*} 
\center
\caption{Results on MARS-Attribute and DukeMTMC-VID-Attribute RGB-Event based PAR dataset.}  
\label{resultsMARSDukeMTMC} 
\begin{tabular}{l|l|cccc|cccc}
\hline \toprule [0.5 pt] 
\multicolumn{1}{c|}{\multirow{2}{*}{\textbf{Methods}}} & \multicolumn{1}{c|}{\multirow{2}{*}{\textbf{Backbone}}} & \multicolumn{4}{c|}{\textbf{MARS-Attribute Dataset}} & \multicolumn{4}{c}{\textbf{DukeMTMC-VID-Attribute Dataset}}  \\ \cline{3-10} 
\multicolumn{1}{c|}{} &
\multicolumn{1}{c|}{} &
\multicolumn{1}{c}{\textbf{Acc}} &
\multicolumn{1}{c}{\textbf{Prec}} &
\multicolumn{1}{c}{\textbf{Recall}} &
\multicolumn{1}{c|}{\textbf{F1}} &
\multicolumn{1}{c}{\textbf{Acc}} &
\multicolumn{1}{c}{\textbf{Pren}} &
\multicolumn{1}{c}{\textbf{Recall}} &
\multicolumn{1}{c}{\textbf{F1}} \\  
\hline
\cellcolor{white}VTB \cite{cheng2022VTB} & ViT-B/16 &72.73   &82.79  & 83.52&  82.89 & 72.50   &83.60   &82.24  &82.38  \\
Zhou et al.~\cite{zhou2023solution} & ConvNext & 69.43  & 81.75 & 80.64 & 81.19  & 65.51   & 78.42  & 77.40 & 77.91 \\
VTB-PLIP ~\cite{zuo2023plip} & ResNet50  & 54.93  & 70.02 &69.23 & 69.14  &43.95    &60.80   & 58.52 & 59.10 \\
Rethink-PLIP~\cite{zuo2023plip} & ResNet50 &  47.85 & 62.15 & 64.80 &  63.45 &  40.25  &  55.41  & 56.13 & 55.77 \\
SSPNet ~\cite{zhou2024pedestrian} & ResNet50& 65.68&77.09  & 79.74 & 78.39  &  67.53  & 78.54  & 79.78 & 79.15 \\
MambaPAR~\cite{wang2024SSMSurvey} & Vim & 69.07  & 81.83 & 79.39 & 80.24  &  61.55  & 75.54  & 74.84 & 74.24 \\
\hline 
OTN-RWKV(Ours) &RWKV-B  &\textbf{73.21}   &\textbf{85.63}  & 81.53&\textbf{83.22} &  \textbf{73.15} &\textbf{84.45}  &82.16  &\textbf{82.78}  \\
\hline \toprule [0.5 pt] 
\end{tabular} 
\end{table*}

\subsection{Implementation Details} 
In our experiments, we use the VRWKV6-B version of the VRWKL~\cite{duan2024vision} base model pre-trained on the ImageNet-1K~\cite{deng2009imagenet} dataset as the visual encoder. This version consists of 12 block layers. We select SGD~\cite{gower2019sgd} as the optimizer, We leverage the warm-up strategy and increase the learning rate from 0 to the initial learning rate 0.008 linearly in the first 10 epochs, and decrease the learning rate by a factor of 0.1 when the number of iterations increases, we set the batch size to 16 and train for 60 epochs, the filtering threshold is set to 0.75. During both the training and inference stages, we first pad and resize the images to 256×128. Training images are augmented with random horizontal flipping with a probability of 0.5 and random cropping with a padding size of 10. Feature interaction is performed using the last layer of the transformer. Our model is implemented based on the PyTorch deep learning framework, and the experiments are conducted on a server equipped with an NVIDIA RTX 3090 GPU. For more details about our framework, please refer to our source code.

\subsection{Comparison on Public Datasets}
In this section, we conduct a detailed comparison between our model and existing pedestrian attribute recognition (PAR) algorithms on three benchmark datasets. Since the MARS-Attribute and DukeMTMC-VID-Attribute datasets contain only the RGB modality, we simulate the corresponding event data to ensure the completeness and fairness of the experiments.

\noindent $\bullet$ \textbf{Results on EventPAR Datasets.~}
We collect and analyze public PAR methods from 2015 to 2025 on the EventPAR dataset as shown in Table~\ref{resultsEventPAR}, methods such as SequencePAR~\cite{jin2023sequencepar}, and PARFormer~\cite{fan2023parformer} have demonstrated superior performance. Our method achieves good results using only RGB data, with mA, Acc, and F1 scores of 79.32, 76.00, and 83.22 respectively. Adding the event modality further improves these to 87.66, 84.78, and 89.07, achieving the best overall performance and demonstrating the significant benefit of event data for model robustness in complex environments. Because PAR methods from 2025 have not released their code, experiments cannot be conducted.

\noindent $\bullet$ \textbf{Results on MARS-Attribute Datasets~\cite{zheng2016mars}.~} 
As shown in Table~\ref{resultsMARSDukeMTMC}, comparing our method with several state-of-the-art approaches. Our method achieves the best performance on most evaluation metrics, achieving 73.21, 85.63, 81.53, and 83.22 in Accuracy, Precision, Recall, and F1, respectively. These results reflect a balanced performance with high accuracy and effective attribute capture.


\noindent $\bullet$ \textbf{Results on DukeMTMC-VID-Attribute Datasets}\textbf{~\cite{ristani2016performance}.}
As shown in Table~\ref{resultsMARSDukeMTMC}, we also compared our method with several mainstream approaches on this dataset, achieving strong results with  accuracy, precision, recall, and F1 scores of 73.15, 84.45, 82.16, and 82.79, respectively. Our method shows stable performance across diverse data environments, confirming its effectiveness and adaptability, with strong application potential.

\subsection{Ablation Study}
In this section, we conduct a comprehensive ablation study on the core modules of EventPAR, including different input configurations, various backbone architectures, fusion strategies and event data aggregation methods. 

\noindent $\bullet$ \textbf{Analysis of different input settings.} 
In this study, We varied RGB input counts and ran extensive experiments as shown in Table~\ref{tab:ablation_inputSetting}. RGB or Event alone performed well, with RGB slightly worse due to noise. Combining both improved results, showing complementarity. Too many RGB inputs hurt performance due to redundancy and noise, burdening feature extraction and weakening semantic focus.

\begin{table}[htbp]
\centering
\small
\caption{Comparison of different input settings.}
\label{tab:ablation_inputSetting}
\resizebox{0.46\textwidth}{!}{ 
\begin{tabular}{c|c|c|c|c|c|c|c}
\hline \toprule[0.5pt]
\multicolumn{1}{c}{\textbf{RGB}} & \multicolumn{1}{c|}{\textbf{Event}} & \multicolumn{1}{c|}{\textbf{mA}} & \multicolumn{1}{c|}{\textbf{Acc}} & \multicolumn{1}{c|}{\textbf{Prec}} & \multicolumn{1}{c|}{\textbf{Recall}} & \multicolumn{1}{c}{\textbf{F1}}  \\ \hline
\multicolumn{1}{c}{1} & \multicolumn{1}{c|}{0} & \multicolumn{1}{c|}{79.41} & \multicolumn{1}{c|}{76.22} & \multicolumn{1}{c|}{82.43} & \multicolumn{1}{c|}{84.65} & \multicolumn{1}{c}{83.27}  \\ 
\multicolumn{1}{c}{0} & \multicolumn{1}{c|}{5} & \multicolumn{1}{c|}{87.14} & \multicolumn{1}{c|}{84.52} & \multicolumn{1}{c|}{89.07} & \multicolumn{1}{c|}{89.09} & \multicolumn{1}{c}{88.91}  \\ 
\multicolumn{1}{c}{1} & \multicolumn{1}{c|}{5} & \multicolumn{1}{c|}{\textbf{87.70}} & \multicolumn{1}{c|}{\textbf{84.94}} & \multicolumn{1}{c|}{\textbf{89.15}} & \multicolumn{1}{c|}{\textbf{89.48}} & \multicolumn{1}{c}{\textbf{89.18}}  \\ 
\multicolumn{1}{c}{3} & \multicolumn{1}{c|}{5} & \multicolumn{1}{c|}{85.97} & \multicolumn{1}{c|}{81.91} & \multicolumn{1}{c|}{86.84} & \multicolumn{1}{c|}{86.98} & \multicolumn{1}{c}{86.61}  \\ 
\multicolumn{1}{c}{5} & \multicolumn{1}{c|}{5} & \multicolumn{1}{c|}{85.69} & \multicolumn{1}{c|}{82.24} & \multicolumn{1}{c|}{87.19} & \multicolumn{1}{c|}{86.88} & \multicolumn{1}{c}{86.75} \\ 
\hline \toprule[0.5pt] 
\end{tabular} }
\end{table}

\begin{table}[htbp]
\centering
\small
\caption{Comparison of different feature fusion methods.}
\label{tab:ablation_fusion}
\resizebox{0.46\textwidth}{!}{ 
\begin{tabular}{c|c|c|c|c|c|c|c}
\hline \toprule[0.5pt]
\multicolumn{1}{c|}{\textbf{Method}}  & \multicolumn{1}{c|}{\textbf{mA}} & \multicolumn{1}{c|}{\textbf{Acc}} & \multicolumn{1}{c|}{\textbf{Prec}} & \multicolumn{1}{c|}{\textbf{Recall}} & \multicolumn{1}{c}{\textbf{F1}}  \\ 
\hline
 \multicolumn{1}{c|}{Concat} & \multicolumn{1}{c|}{87.63} & \multicolumn{1}{c|}{84.60} & \multicolumn{1}{c|}{88.80} & \multicolumn{1}{c|}{89.40} & \multicolumn{1}{c}{88.96}  \\ 
 \multicolumn{1}{c|}{Add} & \multicolumn{1}{c|}{87.17} & \multicolumn{1}{c|}{84.62} & \multicolumn{1}{c|}{88.96} & \multicolumn{1}{c|}{89.30} & \multicolumn{1}{c}{88.99}   \\ 
 \multicolumn{1}{c|}{$1\times1$ Conv} & \multicolumn{1}{c|}{83.77} & \multicolumn{1}{c|}{80.33} & \multicolumn{1}{c|}{84.78} & \multicolumn{1}{c|}{88.63} & \multicolumn{1}{c}{86.44}  \\  
\multicolumn{1}{c|}{OTN-RWKV} & \multicolumn{1}{c|}{\textbf{87.70} }& \multicolumn{1}{c|}{\textbf{84.94}} & \multicolumn{1}{c|}{\textbf{89.15}} & \multicolumn{1}{c|}{\textbf{89.48}} & \multicolumn{1}{c}{\textbf{89.18}}  \\ 
\hline \toprule[0.5pt] 
\end{tabular} }
\end{table}

\begin{table}[htbp]
\centering
\small
\caption{Comparison of different aggregation strategies.}
\label{tab:ablation_aggregation}
\resizebox{0.46\textwidth}{!}{ 
\begin{tabular}{c|cccc|c|c|c}
\hline \toprule[0.5pt]
\multicolumn{1}{c|}{\textbf{No.}} & \multicolumn{1}{c}{\textbf{Max}} & \multicolumn{1}{c}{\textbf{Mean}} & \multicolumn{1}{c}{\textbf{GNN}} & \multicolumn{1}{c|}{\textbf{Sim}} & \multicolumn{1}{c|}{\textbf{mA}} & \multicolumn{1}{c|}{\textbf{Acc}} & \multicolumn{1}{c}{\textbf{F1}} \\
\hline
01 & \checkmark &       &      &      & 87.56 & 84.62 & 88.98 \\
02 &           & \checkmark &      &      & 87.50 & 84.68 & 89.01 \\
03 &           &       & \checkmark &      & 84.53 & 82.24 & 87.50 \\
04 &           &       &      & \checkmark & \textbf{87.59} & \textbf{84.80} & \textbf{89.10} \\
\hline \toprule[0.5pt]
\end{tabular}
}
\end{table}

\noindent $\bullet$ \textbf{Analysis of different feature fusion methods.~} 
As shown in Table~\ref{tab:ablation_fusion},
We compared our proposed OTN-RWKV fusion strategy with three common methods: Concat, Add, and 1×1Conv. OTN-RWKV achieved top results with mA 87.70, ACC 84.94, and F1 89.18, outperforming other methods. It effectively fuses RGB and event features, boosting robustness and performance.

\noindent $\bullet$ \textbf{Analysis of different aggregation strategies of Event.~} 
As shown in Table~\ref{tab:ablation_aggregation}, we systematically compared four event data aggregation strategies: Max Pooling, Mean Pooling, a graph-based GNN, and our proposed Similarity-based approach. Our method adaptively fuses event features, preserves temporal cues, suppresses noise, and validates the effectiveness of the aggregation strategy.

\noindent $\bullet$ \textbf{Analysis of different threshold setting in the similarity aggregation strategies~} 
To further validate the impact of threshold settings in the similarity-based filtering strategy, we conducted experiments with varying similarity thresholds under identical conditions. The threshold controls the degree of fusion between event frames, preserving only those with sufficient semantic similarity to reduce redundant or noisy features. As shown in Table~\ref{tab:ablation_threshold}, a well-chosen threshold better preserves discriminative features and enhances performance, while extreme thresholds may impair fusion quality and degrade results.

\begin{table}[htbp] 
\centering
\small
\caption{Comparison of different threshold setting in the feature filtering module.}
\label{tab:ablation_threshold}
\begin{tabular}{c|c|c|c|c|c|c}
\hline \toprule[0.5pt]
\multicolumn{1}{c|}{\textbf{No.}} & \multicolumn{1}{c|}{\textbf{Threshold}} & \multicolumn{1}{c|}{\textbf{mA}} & \multicolumn{1}{c|}{\textbf{Acc}} & \multicolumn{1}{c|}{\textbf{Prec}} & \multicolumn{1}{c|}{\textbf{Recall}} & \multicolumn{1}{c}{\textbf{F1}}  \\ 
\hline
\multicolumn{1}{c|}{01}&\multicolumn{1}{c|}{0.60} & \multicolumn{1}{c|}{87.70} & \multicolumn{1}{c|}{84.14} & \multicolumn{1}{c|}{88.56} & \multicolumn{1}{c|}{89.07} & \multicolumn{1}{c}{88.68}  \\ 
\multicolumn{1}{c|}{02}&
\multicolumn{1}{c|}{0.65} & \multicolumn{1}{c|}{87.55} & \multicolumn{1}{c|}{84.16} & \multicolumn{1}{c|}{88.62} & \multicolumn{1}{c|}{89.09} & \multicolumn{1}{c}{88.70}  \\ 
\multicolumn{1}{c|}{03}&
 \multicolumn{1}{c|}{0.75} & \multicolumn{1}{c|}{87.70} & \multicolumn{1}{c|}{84.94} & \multicolumn{1}{c|}{89.15} & \multicolumn{1}{c|}{89.48} & \multicolumn{1}{c}{89.18}  \\ 
 \multicolumn{1}{c|}{04}&
 \multicolumn{1}{c|}{0.80} & \multicolumn{1}{c|}{87.62} & \multicolumn{1}{c|}{84.17} & \multicolumn{1}{c|}{88.77} & \multicolumn{1}{c|}{88.94} & \multicolumn{1}{c}{88.71}  \\ 

 \multicolumn{1}{c|}{05}&
 \multicolumn{1}{c|}{0.95} & \multicolumn{1}{c|}{86.53} & \multicolumn{1}{c|}{82.46} & \multicolumn{1}{c|}{87.37} & \multicolumn{1}{c|}{87.49} & \multicolumn{1}{c}{87.12}  \\ 
\hline \toprule[0.5pt] 
\end{tabular} 
\end{table}

\noindent $\bullet$ \textbf{Analysis of different backbone.~}
To validate the effectiveness of the RWKV, we conducted comparisons with ViT and ResNet-50 under the same settings. RWKV achieved superior performance with mA, Accuracy, and F1 of 87.70, 84.94, and 89.18, respectively, demonstrating its advantage in pedestrian attribute recognition, as shown in Table~\ref{tab:ablation_backbone}.
\begin{table}[htbp]
\centering
\renewcommand{\arraystretch}{0.9}
\small 
\caption{Comparison of different backbones.}
\label{tab:ablation_backbone}
\begin{tabular}{c|c|c|c|c}
\hline \toprule[0.5pt]
\multicolumn{1}{c|}{\textbf{No.}} & \multicolumn{1}{c|}{\textbf{Backbone}} & \multicolumn{1}{c|}{\textbf{mA}} & \multicolumn{1}{c|}{\textbf{Acc}} & \multicolumn{1}{c}{\textbf{F1}} \\
\hline
01 & ViT      & 84.05 & 82.41 & 87.60 \\ 
02 & ResNet50 & 83.85 & 82.45 & 87.64 \\
03 & RWKV     & \textbf{87.70} & \textbf{84.94} & \textbf{89.18} \\
\hline \toprule[0.5pt]
\end{tabular}
\end{table}

\subsection{Visualization}
\begin{figure}
    \centering
    \includegraphics[width=\linewidth]{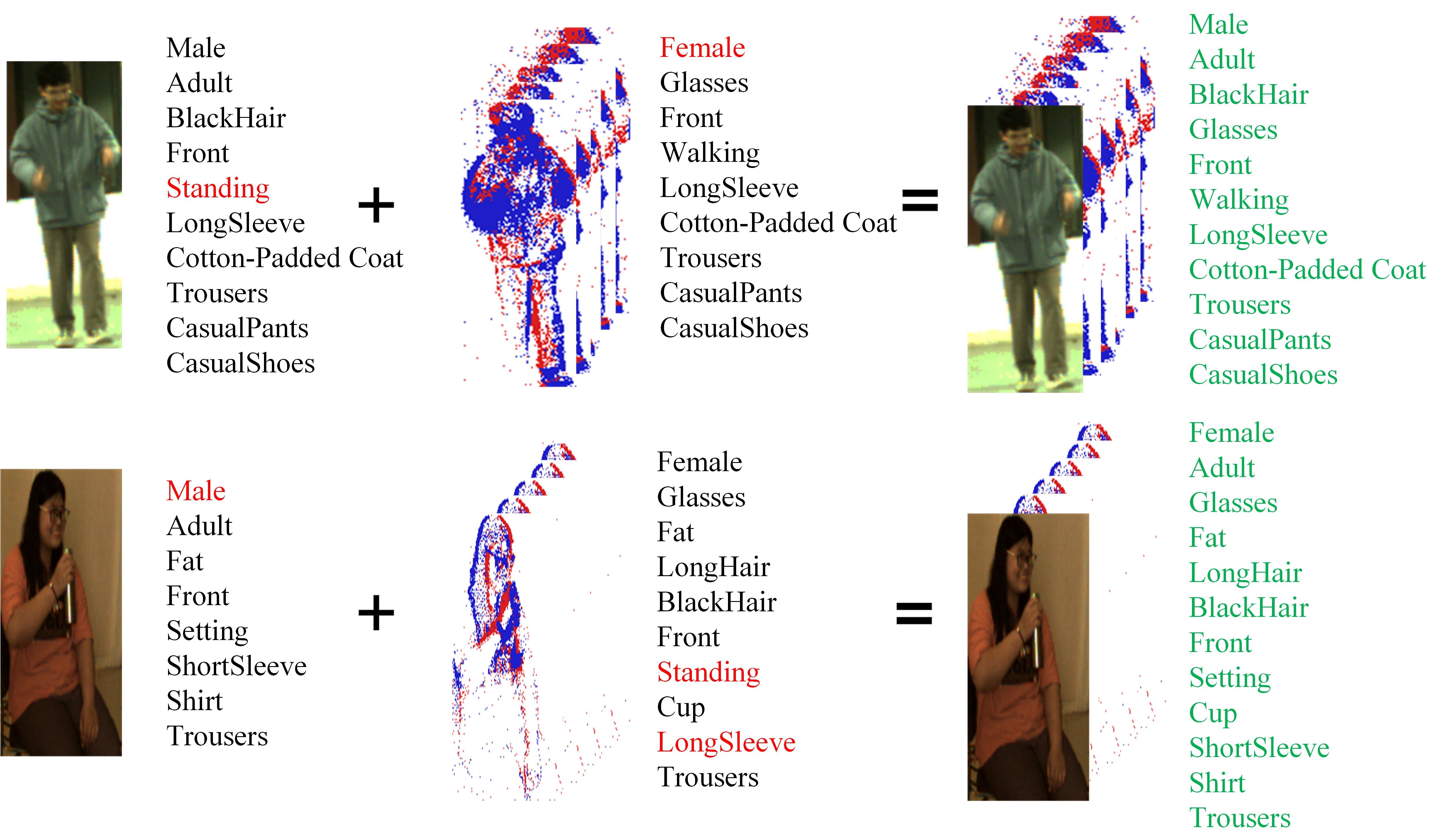}
    \caption{Visualization of pedestrian attributes predicted by our proposed method. The \textcolor{red}{red} attributes indicate incorrect predictions, while \textcolor{green}{green} attributes represent ground truth. } 
    \label{fig:attr_v}
\end{figure}

\noindent $\bullet$ \textbf{Visualization of Predicted Results.~} 
As shown in Fig. \ref{fig:attr_v}, predictions using RGB, event, and fused bimodal inputs indicate that single modalities struggle to provide accurate and comprehensive recognition, while bimodal fusion better captures complementary information, significantly improving accuracy and robustness.

\begin{figure}
    \centering
    \includegraphics[width=\linewidth]{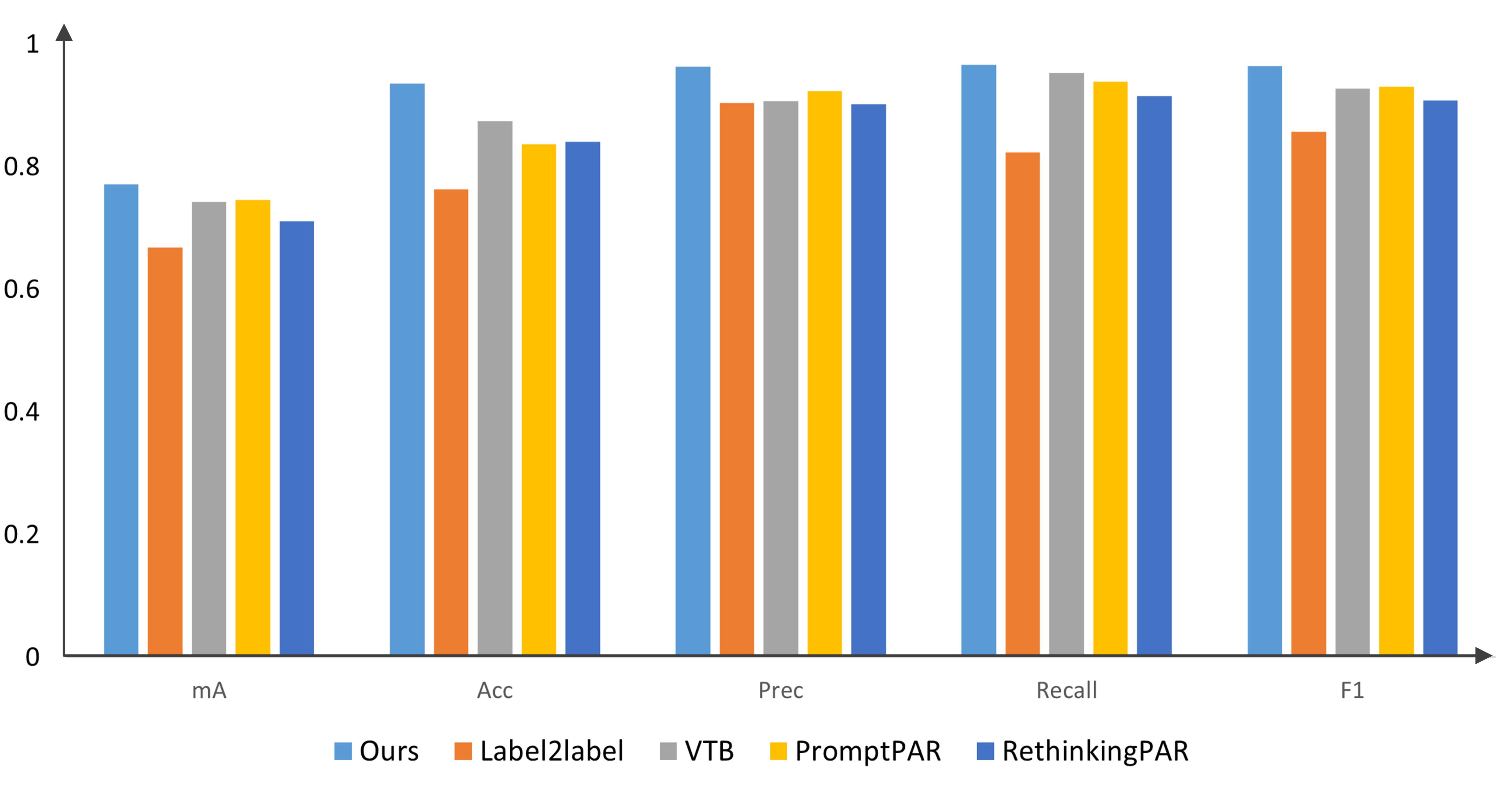}
    \caption{Visualization of emotion attribute prediction results by our method and other compared approaches. } 
    \label{fig:emotion_table}
\end{figure}

\noindent $\bullet$ \textbf{Visualization of emotion attribute prediction.~} 
Fig. \ref{fig:emotion_table} visualizes multiple metrics for six emotion attributes. Our innovative RGB-Event fusion effectively captures emotional cues, consistently outperforming others and validating the value of emotion attributes in pedestrian recognition.

\begin{figure}
    \centering
    \includegraphics[width=\linewidth]{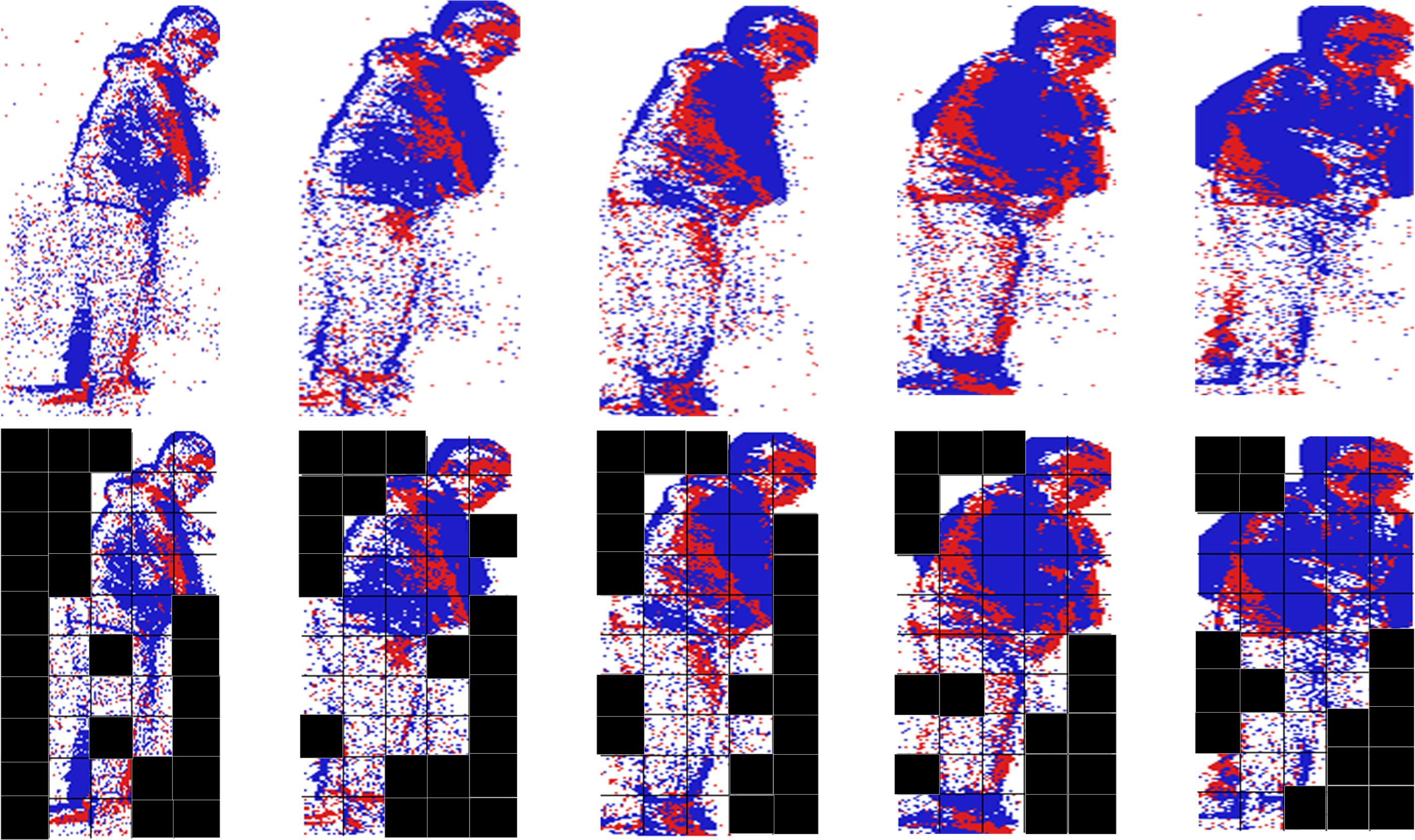}
    \caption{Illustration of our proposed similarity-based aggregation strategy for event token selection, where the black parts indicate the filtered-out tokens.} 
    \label{fig:token_select}
\end{figure}

\noindent $\bullet$ \textbf{Visualization of Aggregation Strategy.~}  
To provide an intuitive understanding of our proposed similarity-based aggregation strategy, we present a visual illustration in Fig. \ref{fig:token_select}. In this figure, the colored regions represent event tokens that are retained and contribute to the final aggregation, while the black areas indicate tokens filtered out by the similarity-based selection process. As shown, the strategy effectively suppresses redundant or irrelevant information by adaptively retaining only semantically relevant tokens. This helps the model focus on discriminative patterns, thereby improving the overall recognition performance.

\section{Conclusion and Future Works}
\label{conclusion}
In this paper, we addressed the limitations of existing pedestrian attribute recognition (PAR) methods, which are predominantly based on RGB frame cameras and suffer from challenges such as sensitivity to lighting conditions and motion blur. Moreover, current approaches focus primarily on external appearance and clothing attributes, neglecting the emotional dimensions of pedestrians. To overcome these constraints, we introduced a novel multi-modal RGB-Event attribute recognition task, leveraging the advantages of event cameras in low-light, high-speed, and low-power scenarios. Our key contribution is the creation of EventPAR, the first large-scale multi-modal PAR dataset, which includes 100K paired RGB-Event samples covering 50 attributes related to both appearance and six human emotions. By retraining and evaluating mainstream PAR models on EventPAR, we established a comprehensive benchmark, offering valuable data and algorithmic baselines for the community. Furthermore, we proposed a novel RWKV-based multi-modal PAR framework, incorporating an RWKV visual encoder and an asymmetric RWKV fusion module. Extensive experiments conducted on EventPAR and two simulated datasets demonstrated state-of-the-art performance, underscoring the effectiveness of our approach. 
In our future works, we will focus on exploiting learning-based new event representations for the perception of event streams for pedestrian attribute recognition.

\small
\bibliographystyle{ieeenat_fullname}
\bibliography{reference}


\end{document}